\documentclass[conference]{IEEEtran}
\IEEEoverridecommandlockouts
\usepackage{cite}
\usepackage{amsmath,amssymb,amsfonts}
\usepackage{algorithmic}
\usepackage{multirow}
\usepackage{graphicx}
\usepackage{textcomp}
\usepackage{hyperref}
\usepackage{xcolor}
\usepackage[most]{tcolorbox}
\usepackage{booktabs}
\def\BibTeX{{\rm B\kern-.05em{\sc i\kern-.025em b}\kern-.08em
    T\kern-.1667em\lower.7ex\hbox{E}\kern-.125emX}}
\begin{document}

\title{LITE: Look at the Image and Trust the Expert}

\author{\IEEEauthorblockN{Alessandro Di Chiara}
\IEEEauthorblockA{\textit{Sapienza University of Rome} \\
Rome, Italy \\
dichiara.1938462@studenti.uniroma1.it }
\and
\IEEEauthorblockN{Simone Teglia}
\IEEEauthorblockA{\textit{Sapienza University of Rome} \\
Rome, Italy \\
teglia@diag.uniroma1.it}
\and
\IEEEauthorblockN{Irene Amerini}
\IEEEauthorblockA{\textit{Sapienza University of Rome} \\
Rome, Italy \\
amerini@diag.uniroma1.it}
\and

}

\maketitle

\begin{abstract} 
% The rapid advancement of generative artificial intelligence has enabled the creation of highly realistic synthetic images that are increasingly indistinguishable from authentic images.
% The rapid proliferation of deepfakes has introduced significant risks in sensitive domains such as political communication, armed conflicts, and public health. Consequently, developing reliable methods for detecting AI-generated images under real-world conditions has become increasingly important. 
Nowadays, Deepfake Detection in real-world conditions is still a challenging task as online content is frequently subject to compression, resizing, repeated sharing, and re-digitization. 
Such transformations can alter or suppress the visual artifacts on which detectors rely, substantially degrading the performance of models trained on clean and controlled datasets.
Motivated by this issue, in this work we first perform a systematic zero-shot benchmark of six VLMs, evaluating different types of prompts, across three different acquisition condition: original, social-media transferred, and re-digitized images. 
Based on this analysis we propose LITE (Look at the Image and Trust the Expert), a condition-aware framework that first estimates the condition of the input image and then routes it to the corresponding condition-specific VLM expert. 
Evaluated on B-RRDataset, LITE achieves 0.967 accuracy and a 0.966 F1-score, surpassing state-of-the-art deepfake detectors trained under the same protocol.
Furthermore, LITE achieves state-of-the-art performance on SID-Set and exhibits strong cross-dataset generalization on TrueFake, despite being trained exclusively on B-RRDataset. These results show that exploiting condition-specific specialization within lightweight VLMs provides an effective strategy for robust deepfake detection under realistic image degradations. We publicly release our code at \url{https://github.com/ALCOR-Lab-DIAG/LITE}.

\end{abstract}

\section{Introduction}

With the rapid advancements of Generative Artificial Intelligence, the quality and accessibility of AI-generated visual content have significantly improved. State-of-the-art models such as DALL-E 3~\cite{betker2023dalle3}, FLUX.1~\cite{flux2024}, and Nano Banana Pro~\cite{nanobananapro2025} can now generate highly realistic images, commonly known as deepfakes.
Although these technologies offer new opportunities for content creation, they have also enabled the misuse of deepfakes for misinformation, political manipulation, and fabricated visual evidence, posing serious risks to public trust and information integrity~\cite{dhiman2023exploding,plada2025,ching2025}. 
Consequently, reliable detection of AI-generated images has become increasingly important. However, real-world deepfake detection remains challenging, as social media compression suppresses forensic traces~\cite{emulatore2025}, while re-digitization processes, such as screen recapture, introduce additional artifacts including geometric distortions, color changes, and resolution loss~\cite{rrdataset2025}.
Current deepfake detectors mainly rely on CNNs and Vision Transformers to learn visual features~\cite{deepforensics2025,xue2022transformer,tan2023lgrad}. Although recent architectures have improved generalization, their performance still degrades under distribution shifts~\cite{moeffd2025,lai2024gmdf, yang2025d3}. Vision-Language Models (VLMs) have recently emerged as a promising alternative due to their strong visual reasoning and zero-shot capabilities, motivating their adoption for deepfake detection~\cite{pirogov2025vlm,canvlm2025,yu2025unlocking,khaliq2026survey}. 
Nevertheless, recent studies indicate that such powerful models remain vulnerable to real-world image transformations, suggesting that their reasoning capabilities alone are insufficient to ensure robust detection in real-world scenarios~\cite{fitforpurpose2025,rrdataset2025}. However, it remains unexplored whether different VLMs exhibit complementary robustness to different types of image degradation and whether such complementarity can be exploited to improve detection performance. 
Motivated by this question, we first conduct an extensive benchmarking of lightweight VLMs on B-RRDataset, a balanced dataset derived from RRDataset~\cite{rrdataset2025}. Because the original RRDataset training split contains only pristine images, we construct B-RRDataset by sampling equal numbers of real and fake images across the original, social-media transfer, and re-digitization conditions. This balanced design enables an unbiased comparison of model robustness under different acquisition conditions. Later, based on the VLM benchmarking, we propose LITE (\textbf{L}ook at the \textbf{I}mage and \textbf{T}rust the \textbf{E}xperts), a condition-aware deepfake detection framework that combines a condition detector with three condition-specialized VLM experts.
We evaluate LITE on B-RRDataset and study its generalization on the SID-Set ~\cite{huang2025sida} and TrueFake~\cite{truefake2025} benchmarks.

In summary, our main contributions are:

\begin{itemize}
    \item A comprehensive analysis of lightweight Vision-Language Models under realistic image degradations, revealing complementary robustness patterns across pristine, social-media-transferred, and re-digitized images.

    \item LITE (\textbf{L}ook at the \textbf{I}mage and \textbf{T}rust the \textbf{E}xpert), a condition-aware deepfake detection framework that dynamically routes each image to a specialized VLM expert according to its acquisition condition.

    \item Extensive experiments demonstrating that LITE surpasses six state-of-the-art deepfake detectors under a unified protocol and achieves strong cross-dataset generalization on SID-Set and TrueFake without additional fine-tuning.
\end{itemize}

\section{Related Works}

\subsection{Deepfake Detection in the Wild}

Early deepfake detectors relied on handcrafted forensic cues, such as
eye-blinking patterns and face-warping artifacts~\cite{li2018eyeblink,
li2019facewarp}. As generative models improved, CNN-based methods became the standard by learning spatial and frequency artifacts directly from data~\cite{frank2020leveraging, mesonet2018, zhong2023patchcraft}, supported by large-scale benchmark datasets such as FaceForensics++~\cite{rossler2019}, Celeb-DF~\cite{li2020celebdf}, and DFDC~\cite{dolhansky2020dfdc}. 
More recently, Vision Transformers have been adopted as powerful backbones for their ability to capture
long-range dependencies~\cite{ wang2022m2tr, zhuang2022uiavit}, while recent methods such as MoE-FFD~\cite{moeffd2025}, CAST~\cite{cast2025}, and UnivFD~\cite{ojha2023univfd} have improved generalization through better adaptation and feature representations. However, most approaches are still evaluated under controlled settings, with limited analysis of robustness under realistic image degradation.
Recent studies have shown that detector performance drops markedly after social-media compression and other post-processing operations~\cite{fitforpurpose2025}, motivating robustness-oriented methods such as PLADA~\cite{plada2025}, SNVSE~\cite{emulatore2025}, and UMCL~\cite{umcl2025}.
In parallel, benchmarks including TrueFake~\cite{truefake2025} and RRDataset~\cite{rrdataset2025} have been introduced to evaluate robustness under social-media transmission and multiple acquisition conditions. Despite these efforts, robust deepfake detection across diverse real-world image conditions remains an open challenge.

\subsection{Vision-Language Models for Deepfake Detection}

Recent Vision-Language Models (VLMs) have emerged as a new paradigm for
visual understanding by combining visual representations with language
reasoning~\cite{radford2021clip,li2023blip2,dai2023instructblip,
liu2023llava,qwen2025vl,internvl2024}. These capabilities motivated the
application of VLMs to deepfake detection, initially formulated as a visual
question answering (VQA) task. AntifakePrompt~\cite{chang2023antifake},
DD-VQA~\cite{zhang2024ddvqa}, FakeBench~\cite{li2024fakebench}, and
SHIELD~\cite{shi2024shield} explored prompt-based detection and visual
explanations. Recent studies showed promising zero-shot performance, although
results remain strongly dependent on the underlying VLM and the adopted  prompting strategy~\cite{jia2024chatgpt,pirogov2025vlm,canvlm2025}.
Nevertheless, existing evaluations are predominantly conducted on pristine datasets or under generic zero-shot settings, providing limited insight into the robustness of VLMs under realistic image degradations. As shown by RRDataset~\cite{rrdataset2025}, social-media transmission and re-digitization can substantially impair VLM performance, indicating that current models lack consistent robustness across different acquisition conditions and motivating the need for approaches specifically designed to address this variability.
Existing methods, including Mixture-of-Experts detectors such as MoE-FFD \cite{moeffd2025}, rely on a single network whose experts or gating are learned end-to-end. LITE instead estimates the acquisition condition explicitly and routes each image to a separate VLM fine-tuned on that condition, making the routing interpretable and the router evaluable on its own.

\section{Methodology}
\label{sec:methodology}
% Current Vision-Language Models (VLMs) have demonstrated promising zero-shot capabilities for deepfake detection, yet their robustness under realistic image degradations remains largely unexplored. Motivated by this, we investigate whether different lightweight VLMs exhibit complementary strengths across different acquisition conditions and whether such complementarity can be exploited to improve deepfake detection.

To investigate whether different lightweight VLMs exhibit complementary strengths across different acquisition conditions, our methodology consists of two stages. First, we systematically benchmark several lightweight open-source VLMs under three image conditions to reveal their relative strengths. The resulting analysis is then used to instantiate LITE, a condition-aware framework that dynamically routes each image to the most suitable specialized VLM expert according to its estimated acquisition condition.

\begin{figure}[!ht]
\centering
\includegraphics[width=\linewidth]{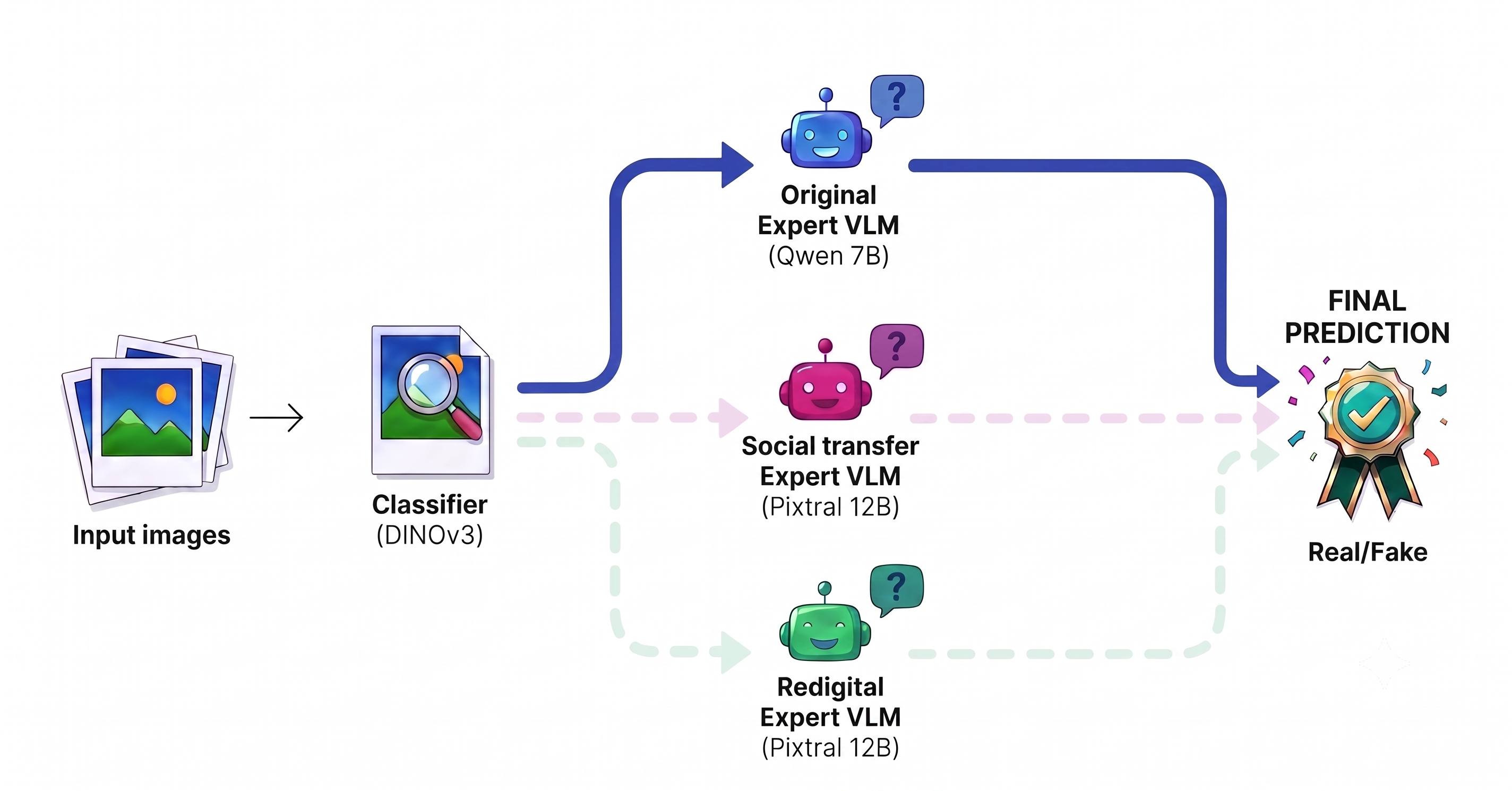}
\caption{Overview of the LITE architecture. The highlighted arrow illustrates the routing path selected by the classifier when the input image is identified as \emph{original}, activating the Original Expert VLM.}
\label{fig:architecture}
\end{figure}

\subsection{Benchmarking VLMs Across Image Conditions}
Before applying any task-specific fine-tuning, we evaluate several lightweight VLMs in a zero-shot setting to assess their capability to distinguish between real and AI-generated images under different acquisition conditions. 
Specifically, we benchmark Qwen2.5-VL-7B-Instruct~\cite{qwen2025vl} \footnote{\url{https://huggingface.co/Qwen/Qwen2.5-VL-7B-Instruct}}, InternVL2.5-8B~\cite{internvl2024}\footnote{\url{https://huggingface.co/OpenGVLab/InternVL2_5-8B}}, Pixtral-12B-2409~\cite{pixtral2024} \footnote{\url{https://huggingface.co/mistralai/Pixtral-12B-2409}}, LLaVA-OneVision-1.5-8B-Instruct~\cite{llavaonevision2025} \footnote{\url{https://huggingface.co/lmms-lab/LLaVA-OneVision-1.5-8B-Instruct}}, PaliGemma2-10B-mix-224~\cite{paligemma2024} \footnote{\url{https://huggingface.co/google/paligemma2-10b-mix-224}}, and DeepSeek-VL-7B~\cite{deepseekvl2024} \footnote{\url{https://huggingface.co/deepseek-ai/deepseek-vl-7b-base}}.
Each model is assessed using two prompting strategies, shown in Figure 2. The first employs an unstructured binary prompt asking whether the image is real or fake. The second uses a structured prompt that guides the model to inspect forensic cues, including texture, lighting, structural consistency, and compression artifacts, before producing the final prediction. The objective of this first phase is twofold. First, it provides a comprehensive assessment of the zero-shot deepfake detection capabilities of lightweight open-source VLMs under realistic image degradations. Second, it identifies the most suitable VLM for each acquisition condition, providing the basis for instantiating the condition-specific experts of LITE. The benchmark reveals that different VLMs excel under different acquisition conditions rather than exhibiting uniformly superior performance. This observation motivates the design of LITE, which leverages condition-specific specialization instead of relying on a single universal detector.

\begin{figure}[!h]
\label{fig:prompts}
\caption{Unstructured and Structured prompts used for the zero-shot evaluation. Prompts have been condensed for presentation. Core criteria and instruction remain identical.}

\begin{tcolorbox}[
    colback=blue!5!white,
    colframe=blue!60!black,
    arc=1mm,
    boxrule=0.5pt,
    left=1pt, right=1pt, top=2pt, bottom=2pt, % Minimal lateral and vertical padding
    fonttitle=\bfseries\small,
    title=Zero-Shot Prompts
]
\fontsize{7.5pt}{9pt}\selectfont % Smaller, readable text with tight line spacing

\textbf{Unstructured Prompt:} \textit{Determine if this image is a real photograph or AI-generated considering visual inconsistencies. Output REAL or FAKE only without explanation.}

\vspace{2pt}
\textbf{Structured Prompt:} \textit{Analyze the image criteria: \textbf{1) Texture:} Smoothness, over-sharpening, or noise artifacts. \textbf{2) Lighting:} Directional consistency, shadows, and reflection coherence. \textbf{3) Coherence:} Physical plausibility, distorted shapes, or geometry flaws. \textbf{4) Integrity:} Repeating patterns, warped regions, or text corruption. \textbf{5) Degradation:} Compression vs. synthetic artifacts. \textbf{6) Realism:} Natural appearance. Output REAL (if real) or FAKE (if AI-generated) only.}
\end{tcolorbox}
\end{figure}

\subsection{LITE Architecture}
Based on the outcome of the benchmarking stage, we propose LITE (Figure \ref{fig:architecture}), a condition-aware architecture composed of an Image-Condition Classifier and three specialized VLM experts. 
Let
\(
\mathcal{X}
\)
denote the space of input images and
\(
\mathcal{Y}=\{0,1\}
\)
the binary label space, where 0 and 1 indicate real and AI-generated images, respectively.
Each image also belongs to one acquisition condition

\[
c\in\mathcal{C}=
\{\textit{original},
\textit{transmitted},
\textit{re-digitized}\},
\]

which reflects the degradation process undergone by the image.
Given an input image $x \in \mathcal{X}$, the Image-Condition Classifier estimates the probability distribution
\[
P(c|x)
=
[p_1,p_2,p_3],
\]
over the three acquisition conditions. The condition with maximum probability is selected,
\[
\hat c
=
\arg\max_c
P(c|x),
\]
and the corresponding expert $ f_c: \mathcal{X} \to \mathcal{Y}$ is activated to perform the final authenticity prediction
\[
\hat y = f_{\hat c}(x)
\]

This routing strategy follows the central principle of LITE (\textbf{L}ook at the \textbf{I}mage and \textbf{T}rust the \textbf{E}xpert): rather than averaging or combining the predictions of multiple experts, only the expert specialized for the estimated acquisition condition processes the image. Besides reducing computational cost during inference, this design allows every expert to learn representations tailored to the forensic characteristics of a single degradation scenario.
Each expert is initialized from the candidate VLM that achieved the highest zero-shot performance on its corresponding acquisition condition during the benchmarking stage. It is then fine-tuned exclusively using training images from that condition, enabling condition-specific specialization.

\section{Experimental Setup}

\subsection{Dataset}
We evaluate LITE on three complementary benchmarks. RRDataset serves as the primary benchmark for zero-shot analysis, expert fine-tuning, and comparison with classic detectors. SID-Set and TrueFake are used to assess cross-dataset generalization, as they are publicly available benchmarks of fully synthetic images shared through real social-media platforms, which is the real-world scenario targeted by LITE.

\subsubsection{RRDataset} This benchmark is specifically designed to evaluate deepfake detectors under three image conditions: \emph{original}, containing pristine images; \emph{social media transfer}, containing images shared through social-media platforms for multiple rounds; and \emph{re-digitization}, comprising images physically recaptured from screens.
To ensure a balanced evaluation across image conditions and classes, we construct a balanced subset of RRDataset, hereafter referred to as \textbf{B-RRDataset}. This subset contains 24K images equally distributed across the three acquisition conditions and the two classes (real and fake). Of these, 18K are used for training and validation, while the remaining 6K are reserved for the evaluation. The same B-RRDataset partition is adopted for the zero-shot benchmark, LITE training, and all state-of-the-art detectors, ensuring a fair comparison under a common evaluation protocol.

\subsubsection{SID-Set} This dataset is explicitly designed to test detector resilience against aggressive post-processing algorithms used by social-media platforms. It comprises 300K images across three different categories: \emph{Real}, \emph{Fully-Synthetic} and \emph{Tampered}.
Because LITE is specifically designed for full-image generative deepfake detection rather than localized forgery manipulation, we assess its generalization capabilities on the official test split using only the \emph{Real} and \emph{Fully Synthetic} categories.

\subsubsection{TrueFake} This large-scale benchmark contains 600K images, comprising high-quality real images and AI-Generated images by GAN and Diffusion Models. Moreover, TrueFake includes 180K images shared only once on three social-media platforms: Facebook, Telegram and X. Our architecture is evaluated on the complete set of social-media-shared images.

\subsection{LITE Instantiation}

% \subsubsection{Candidate VLMs}
% To construct a comprehensive study across different image conditions we evaluate six recent lightweight open-source VLMs that can be locally deployed and fine-tuned, making them suitable for our condition-specific specialization framework. Specifically we test 
% Qwen2.5-VL-7B-Instruct~\cite{qwen2025vl},
% InternVL2.5-8B~\cite{internvl2024},
% Pixtral-12B-2409~\cite{pixtral2024},
% LLaVA-OneVision-1.5-8B-Instruct~\cite{llavaonevision2025},
% PaliGemma2-10B-mix-224~\cite{paligemma2024},and
% DeepSeek-VL-7B~\cite{deepseekvl2024}.

\subsubsection{Image-Condition Classifier}
The Image-Condition Classifier is the initial routing module of LITE and estimates the probability that an input image belongs to each acquisition condition.
To estimate the condition of each image on our primary evaluation benchmark we employ the vision-encoder DINOv3~\cite{simeoni2025dinov3}, specifically \texttt{facebook/dinov3-vitl16-pretrain-lvd1689m} model.
The image-condition classifier is trained using the Linear-Probing followed by Fine-Tuning (LP-FT) protocol~\cite{kumar2022finetuning}. A linear classification head is first optimized while keeping the DINOv3 backbone frozen. Subsequently, only the last four transformer blocks together and the classification head are fine-tuned.

\subsubsection{Condition-Specific experts}
Following the results of the zero-shot benchmark, we select the highest-performing VLM for each acquisition condition as corresponding expert.
Specifically, Qwen2.5-VL-7B is selected as expert for original images while Pixtral-12B is selected for the two remaining conditions. Each selected model is then independently fine-tuned using only the training images corresponding to its assigned acquisition condition.
We fine-tune the experts using Low-Rank Adaptation (LoRA)~\cite{hu2022lora} with the same structured prompt adopted for the zero-shot evaluation. \footnote{Further implementation details can be found in the official \href{https://github.com/ALCOR-Lab-DIAG/LITE}{repository}.}

\section{Results}
% In this section, we present the experimental evaluation of the proposed
% approach. We first analyze the zero-shot capabilities of different VLMs to
% identify the most suitable model for each acquisition condition. We
% then evaluate the complete LITE pipeline by comparing it with classic
% deepfake detectors and finally study its generalization ability on independent
% real-world benchmarks.
This section evaluates the zero-shot performance of candidate VLMs, assesses the complete LITE pipeline against traditional deepfake detectors, and tests its generalization across independent benchmarks.

\begin{table}[!h]
\caption{Zero-shot accuracy under unstructured (U) and structured (S) prompting on B-RRDataset. Values in \textbf{bold} highlight the best score for each model between the two prompting strategies. Values in \colorbox{blue!25}{lilac} highlight the best accuracy on the three image-condition splits.}
\label{tab:zeroshot}
\resizebox{\columnwidth}{!}{%
\small
\setlength{\tabcolsep}{2pt}
\begin{tabular}{lcccccccc}
\toprule
& \multicolumn{2}{c}{Original}
& \multicolumn{2}{c}{Social Transfer}
& \multicolumn{2}{c}{Re-digitization}
& \multicolumn{2}{c}{Overall} \\
\cmidrule(lr){2-3}
\cmidrule(lr){4-5}
\cmidrule(lr){6-7}
\cmidrule(lr){8-9}
Model & U & S & U & S & U & S & U & S \\
\midrule
PaliGemma2-10B~\cite{paligemma2024}
& 0.490 & \textbf{0.501}
& 0.413 & \textbf{0.598}
& 0.488 & \textbf{0.497}
& 0.464 & \textbf{0.498} \\
DeepSeek-VL-7B~\cite{deepseekvl2024}
& 0.460 & \textbf{0.530}
& 0.450 & \textbf{0.507}
& 0.433 & \textbf{0.526}
& 0.449 & \textbf{0.522} \\
InternVL2.5-8B~\cite{internvl2024}
& 0.530 & \textbf{0.546}
& 0.543 & \textbf{0.564}
& \textbf{0.521} & 0.509
& 0.531 & \textbf{0.539} \\
LLaVA-OneVision-1.5-8B~\cite{llavaonevision2025}
& 0.521 & \textbf{0.564}
& 0.525 & \textbf{0.571}
& 0.521 & \textbf{0.578}
& 0.522 & \textbf{0.571} \\
Qwen2.5-VL-7B~\cite{qwen2025vl}
& 0.590 & \colorbox{blue!25}{\textbf{0.706}}
& 0.546 & \textbf{0.600}
& 0.527 & \textbf{0.560}
& 0.555 & \textbf{0.624} \\
Pixtral-12B~\cite{pixtral2024}
& 0.598 & \textbf{0.650}
& 0.609 & \colorbox{blue!25}{\textbf{0.652}}
& 0.538 & \colorbox{blue!25}{\textbf{0.581}}
& 0.581 & \textbf{0.627} \\
\bottomrule
\end{tabular}%
}
\end{table}

\subsection{VLMs Zero-Shot Evaluation}

Table~\ref{tab:zeroshot} reports the zero-shot accuracy of the six evaluated VLMs using unstructured and structured prompting across the three image conditions. As shown in the Table, structured prompting improves the performance of almost all models on all types of condition. This shows that providing explicit instructions about relevant forensic aspects, such as texture inconsistencies, lighting artifacts, and structural coherence, helps the VLMs to better analyze synthetic images. 
Under structured prompting, Qwen2.5-VL-7B achieves the highest accuracy on original images, while Pixtral-12B obtains the best results on both social-media-transferred and re-digitized images. Interestingly, even with 3B more parameters than Qwen, PaliGemma2-10B achieves the lowest performance among the evaluated models, remaining close to 0.5 accuracy with both structured and unstructured prompt.
Based on these results, Qwen2.5-VL-7B is selected as the expert model for the original condition, whereas Pixtral-12B is selected for the social-media-transferred and re-digitized conditions.
Although structured prompting improves the zero-shot results, the overall
performance of the VLMs remains limited, demonstrating that reliable deepfake detection still requires target-tailored approaches like LITE's condition-specific fine-tuning.

\begin{table}[!h]
\caption{Comparison between LITE and state-of-the-art deepfake detectors on B-RRDataset.}
\label{tab:lite-vs-baselines-category}
\resizebox{\columnwidth}{!}{%
\small
\setlength{\tabcolsep}{2pt} 
\begin{tabular}{lccccccccccc}
\toprule
 & \multicolumn{3}{c}{Original} & \multicolumn{3}{c}{Social Transfer} & \multicolumn{3}{c}{Re-digitization} & \multicolumn{2}{c}{Overall} \\
\cmidrule(lr){2-4} \cmidrule(lr){5-7} \cmidrule(lr){8-10} \cmidrule(lr){11-12}
Model & Acc. $\uparrow$ & FPR $\downarrow$ & FNR $\downarrow$ & Acc. $\uparrow$ & FPR $\downarrow$ & FNR $\downarrow$ & Acc. $\uparrow$ & FPR $\downarrow$ & FNR $\downarrow$ & Acc. $\uparrow$ & F1 $\uparrow$ \\
\midrule
LGrad~\cite{tan2023lgrad}       & 0.805 & 0.181 & 0.213 & 0.823 & 0.181 & 0.174 & 0.744 & 0.245 & 0.268 & 0.790 & 0.789 \\
GramNet~\cite{liu2020gramnet}   & 0.884 & 0.082 & 0.150 & 0.874 & 0.095 & 0.157 & 0.858 & 0.132 & 0.152 & 0.872 & 0.869 \\
UnivFD~\cite{ojha2023univfd}    & 0.904 & 0.077 & 0.116 & 0.905 & 0.112 & 0.078 & 0.894 & 0.094 & 0.118 & 0.901 & 0.900 \\
Fusing~\cite{ju2022fusing}      & 0.912 & 0.020 & 0.165 & 0.922 & 0.027 & 0.140 & 0.902 & \textbf{0.023} & 0.182 & 0.912 & 0.901 \\
DRCT-ConvB~\cite{chen2024drct} & 0.960 & 0.041 & 0.028 & 0.945 & 0.050 & 0.060 & 0.928 & 0.084 & 0.059 & 0.946 & 0.947 \\
CNNSpot~\cite{wang2020cnnspot} & {0.957} & 0.027 & 0.060 & 0.950 & 0.037 & 0.063 & \textbf{0.956} & 0.051 & \textbf{0.038} & 0.954 & 0.954 \\
\midrule
Qwen2.5-VL-7B & \textbf{0.983} & 0.016 & \textbf{0.019} & 0.975 & 0.027 & \textbf{0.023} & 0.922 & 0.048 & 0.108 & 0.960 & 0.959 \\
\midrule
\textbf{LITE (ours)} & 0.982 & \textbf{0.015} & 0.022 & \textbf{0.976} & \textbf{0.022} & 0.026 & {0.942} & {0.039} & {0.077} & \textbf{0.967} & \textbf{0.966} \\
\bottomrule
\end{tabular}%
}
\end{table}

\subsection{Comparison with State-Of-The-Art Detectors}
To evaluate the effectiveness of condition-specific experts, we compare
LITE against six fine-tuned state-of-the-art deepfake detectors under identical training and evaluation protocol. As reported in Table~\ref{tab:lite-vs-baselines-category}, LITE achieves superior performance across the majority of acquisition conditions. Specifically, on the Original and Social Transfer images, LITE significantly outperforms the strongest baselines, achieving an accuracy of 0.982 and 0.976 respectively, while maintaining exceptionally low False Positive Rates. Notably, the detector correctly identifies the acquisition condition in 90\% of the B-RRDataset test images, routing most inputs to the appropriate expert. Moreover, LITE achieves the best overall performance among all evaluated methods, reaching 0.967 accuracy and 0.966 F1-score. Compared with the strongest baseline methods, LITE outperforms strong deepfake detectors like DRCT-ConvB~\cite{chen2024drct} and CNNSpot~\cite{wang2020cnnspot} by 2.1 and 1.3 accuracy points respectively. 
Furthermore, LITE achieves the best overall trade-off between FPR and FNR among all compared approaches, with values of 0.025 and 0.042. These results indicate that the improvement is not limited to a single metric but reflects a more balanced detection performance. This superior performance confirms the effectiveness of dynamically routing input images to specialized VLM experts, allowing the framework to adapt its forensic reasoning to the distinct degradation characteristics of each input.

%To isolate the contribution of routing, Table~\ref{tab:sota} also reports a single Qwen2.5-VL-7B fine-tuned on all three conditions under the same protocol. This model matches LITE on original and socially transferred images, but on re-digitized images its FNR rises to 0.108 and its accuracy drops to 0.922, against 0.942 for LITE. The advantage of LITE therefore comes from the re-digitization expert, showing that condition-specific specialization is most useful where the acquisition shift is largest.

\begin{table}[!ht]
\centering
\caption{Comparison on SID-Set (Real and Fully Synthetic categories). Results of the reference methods are taken from the SIDA paper~\cite{huang2025sida}, while LITE is evaluated in this work.}
\label{tab:sidset}
\small
\begin{tabular}{lcccc}
\toprule
Model & \multicolumn{2}{c}{Real} & \multicolumn{2}{c}{Fully Synthetic} \\
\cmidrule(lr){2-3} \cmidrule(lr){4-5}
 & Acc. & F1 & Acc. & F1 \\
\midrule
AntifakePrompt~\cite{chang2023antifake} & 64.8 & 78.6 & 93.8 & 96.8 \\
CNNSpot~\cite{wang2020cnnspot}          & 79.8 & 88.7 & 39.5 & 56.6 \\
FreDect~\cite{frank2020leveraging}      & 83.7 & 91.1 & 16.8 & 28.8 \\
Fusing~\cite{ju2022fusing}              & 85.1 & 92.0 & 34.0 & 50.7 \\
GramNet~\cite{liu2020gramnet}           & 70.1 & 82.4 & 93.5 & 96.6 \\
UnivFD~\cite{ojha2023univfd}            & 68.0 & 67.4 & 62.1 & 87.5 \\
LGrad~\cite{tan2023lgrad}               & 64.8 & 78.6 & 83.5 & 91.0 \\
LNP  ~\cite{bi2023}                                  & 71.2 & 83.2 & 91.8 & 95.7 \\
SIDA-7B~\cite{huang2025sida}            & 89.1 & 91.0 & 98.7 & 98.6 \\
SIDA-13B~\cite{huang2025sida}           & 89.6 & 91.1 & 98.5 & 98.7 \\
\midrule
\textbf{LITE (ours)}                    & \textbf{98.6} & \textbf{99.3} & \textbf{99.9} & \textbf{99.3} \\
\bottomrule
\end{tabular}
\end{table}

\subsection{Out-Of-Distribution Evaluation}
To evaluate the generalization ability of LITE beyond B-RRDataset, we test the pipeline without additional fine-tuning on the SID-Set and TrueFake benchmarks. Since both benchmarks do not contain screen-captured images, we drop the re-digitization expert from the architecture: the classifier output for the re-digitization condition is discarded, and each image is routed to whichever of the two remaining experts has the higher probability. In this setting, the condition classifier achieves an accuracy of 89\% on both SID-Set and TrueFake. As shown in Table~\ref{tab:sidset}, LITE outperforms all competing methods across both Real and Fully-Synthetic images, which suggests that the proposed VLM experts do not merely memorize artifacts specific to B-RRDataset. 
Evaluated on social media transmitted images of TrueFake, as detailed on Table \ref{tab:truefake_experiments}, LITE consistently achieves the highest TPR across all evaluated generative families and platforms. On authentic images, however, LITE exhibits a True Negative Rate (TNR) trade-off, depending on the image source. 
While it maintains strong TNR on FORLAB, it experiences a noticeable drop on FFHQ.
To account for this trade-off, we also computed a class-balanced accuracy, averaging the TNR over the two real sources and the TPR over the eight generators. Under this metric LITE obtains 0.720, 0.727 and 0.727 on FB, TL and X, outperforming every baseline.
Possible explanation is that FFHQ images undergo extensive preprocessing, including face alignment, cropping, and aesthetic enhancement, resulting in a visual appearance that partially resembles the polished outputs of modern image generators. This distribution shift may encourage the VLM experts to produce more conservative predictions, increasing false positives on authentic images.

\begin{table}[t]
\caption{TPR and TNR on TrueFake social-media-shared images. All models are fine-tuned on B-RRDataset under the same protocol as LITE. Results are reported separately for each TrueFake subset after sharing through Facebook (FB), Telegram (TL), and X.}
\label{tab:truefake_experiments}
\begin{center}
\resizebox{\columnwidth}{!}{%
\renewcommand{\arraystretch}{0.85}
\setlength{\tabcolsep}{2.5pt}
\footnotesize
\begin{tabular}{llrrrrrrr}
\toprule
\textbf{Data} & & {LGrad} & {GramNet} & {Fusing} & {DRCT} & {CNNSpot} & {UnivFD} & \textbf{LITE} \\
\textbf{Metric} & & & & & {ConvB} & & &  \\
\midrule
\multirow{3}{*}{\shortstack[l]{FFHQ \\ \tiny TNR}}
 & FB & 0.539 & 0.595 & \textbf{0.906} & 0.711 & 0.665 & 0.835 & 0.098 \\
 & TL & 0.548 & 0.595 & \textbf{0.919} & 0.862 & 0.679 & 0.868 & 0.122 \\
 & X  & 0.543 & 0.598 & 0.932 & \textbf{0.944} & 0.675 & 0.859 & 0.083 \\
\midrule
\multirow{3}{*}{\shortstack[l]{FORLAB \\ \tiny TNR}}
 & FB & 0.684 & 0.724 & \textbf{0.965} & 0.819 & 0.846 & 0.714 & 0.815 \\
 & TL & 0.687 & 0.724 &\textbf{ 0.960} & 0.788 & 0.851 & 0.727 & 0.825 \\
 & X  & 0.683 & 0.729 & \textbf{0.970} & 0.940 & 0.850 & 0.722 & 0.848 \\
\midrule\midrule
\multirow{3}{*}{\shortstack[l]{SG \\ \tiny TPR}}
 & FB & 0.526 & 0.486 & 0.164 & 0.365 & 0.497 & 0.504 & \textbf{0.973} \\
 & TL & 0.514 & 0.484 & 0.143 & 0.138 & 0.480 & 0.453 & \textbf{0.970} \\
 & X  & 0.519 & 0.482 & 0.119 & 0.030 & 0.488 & 0.507 & \textbf{0.983} \\
\midrule
\multirow{3}{*}{\shortstack[l]{SG2 \\ \tiny TPR}}
 & FB & 0.534 & 0.573 & 0.181 & 0.309 & 0.525 & 0.371 & \textbf{0.953} \\
 & TL & 0.530 & 0.573 & 0.173 & 0.160 & 0.509 & 0.335 & \textbf{0.953} \\
 & X  & 0.531 & 0.571 & 0.150 & 0.046 & 0.513 & 0.365 & \textbf{0.973} \\
\midrule
\multirow{3}{*}{\shortstack[l]{SG3 \\ \tiny TPR}}
 & FB & 0.516 & 0.593 & 0.139 & 0.497 & 0.487 & 0.477 & \textbf{0.986} \\
 & TL & 0.506 & 0.593 & 0.121 & 0.277 & 0.461 & 0.410 & \textbf{0.977} \\
 & X  & 0.516 & 0.590 & 0.096 & 0.087 & 0.464 & 0.448 & \textbf{0.987} \\
\midrule
\multirow{3}{*}{\shortstack[l]{SD1.5 \\ \tiny TPR}}
 & FB & 0.454 & 0.489 & 0.262 & 0.711 & 0.492 & 0.633 & \textbf{0.990} \\
 & TL & 0.481 & 0.489 & 0.256 & 0.713 & 0.491 & 0.626 & \textbf{0.988} \\
 & X  & 0.456 & 0.488 & 0.288 & 0.819 & 0.488 & 0.599 & \textbf{0.988} \\
\midrule
\multirow{3}{*}{\shortstack[l]{SD2.1 \\ \tiny TPR}}
 & FB & 0.498 & 0.493 & 0.198 & 0.482 & 0.521 & 0.545 & \textbf{0.986} \\
 & TL & 0.513 & 0.491 & 0.198 & 0.447 & 0.532 & 0.569 & \textbf{0.987} \\
 & X  & 0.517 & 0.491 & 0.208 & 0.489 & 0.531 & 0.554 & \textbf{0.989} \\
\midrule
\multirow{3}{*}{\shortstack[l]{SD3 \\ \tiny TPR}}
 & FB & 0.560 & 0.429 & 0.210 & 0.619 & 0.467 & 0.627 & \textbf{0.988} \\
 & TL & 0.566 & 0.432 & 0.214 & 0.644 & 0.457 & 0.627 & \textbf{0.982} \\
 & X  & 0.545 & 0.429 & 0.202 & 0.671 & 0.460 & 0.637 & \textbf{0.991} \\
\midrule
\multirow{3}{*}{\shortstack[l]{SDXL \\ \tiny TPR}}
 & FB & 0.579 & 0.588 & 0.385 & 0.881 & 0.627 & 0.743 & \textbf{0.996} \\
 & TL & 0.564 & 0.587 & 0.396 & 0.894 & 0.618 & 0.731 & \textbf{0.995} \\
 & X  & 0.549 & 0.585 & 0.377 & 0.835 & 0.622 & 0.743 & \textbf{0.996} \\
\midrule
\multirow{3}{*}{\shortstack[l]{FLUX.1 \\ \tiny TPR}}
 & FB & 0.649 & 0.570 & 0.482 & 0.952 & 0.639 & 0.769 & \textbf{0.994} \\
 & TL & 0.647 & 0.572 & 0.494 & 0.968 & 0.633 & 0.777 & \textbf{0.995} \\
 & X  & 0.639 & 0.571 & 0.486 & 0.960 & 0.638 & 0.794 & \textbf{0.998} \\
\bottomrule
\end{tabular}%
}
\end{center}
\end{table}

\section{Conclusion}
We presented LITE, a condition-aware deepfake detection framework that combines an image-condition classifier with three specialized VLM experts. LITE consistently outperforms sompeting methods on B-RRDataset and demonstrate strong generalization on SID-Set and TrueFake. 
Future works will focus on extending LITE to additional real-world degradations as well as investigating multimodal extensions that combine visual information with textual cues.
Another promising direction is the integration of visual attribution techniques to identify the image regions that drive the predictions of each expert.

% By routing each input image to the expert specialized for its acquisition condition, LITE consistently outperforms competing methods on B-RRDataset, achieves state-of-the-art performance on SID-Set, and demonstrates strong cross-dataset generalization on TrueFake without additional fine-tuning.
% These findings indicate that condition-aware specialization is an effective strategy for improving the robustness of lightweight VLM-based deepfake detectors under realistic image degradations.

% As future work, we aim to extend LITE into a more complete framework that can handle a wider range of real-world degradation scenarios beyond the conditions considered in this work. Moreover, we plan to investigate multimodal extensions that combine visual information with textual and contextual cues to improve robustness in challenging real-world scenarios.

\end{document}